%% file: latex/acl_latex.tex
\pdfoutput=1
\documentclass[11pt]{article}

\usepackage[final]{acl}

\usepackage{times}
\usepackage{latexsym}

\usepackage[T1]{fontenc}
\usepackage[utf8]{inputenc}

\usepackage{microtype}

\usepackage{inconsolata}

\usepackage{graphicx}

\usepackage{amsmath}
\usepackage{multicol}
\usepackage{multirow}
\usepackage{booktabs}
\usepackage{tabularx}
\usepackage[table]{xcolor}
\usepackage{booktabs}
\usepackage{float}
\usepackage{longtable}

\usepackage[most]{tcolorbox}
\usepackage{enumitem}
\usepackage{courier} % optional: nicer monospace

\newtcolorbox{promptbox}[2][]{%
  enhanced,
  breakable,
  colback=gray!3,
  colframe=black!40,
  boxrule=0.6pt,
  arc=2mm,
  left=1.2mm,
  right=1.2mm,
  top=1.0mm,
  bottom=1.0mm,
  title=\textbf{#2},
  fonttitle=\normalsize,
  coltitle=black,
  attach title to upper,
  #1
}

\title{NoisyCausal: A Benchmark for Evaluating Causal Reasoning Under Structured Noise}

\author{
 \textbf{Zhi Xu\textsuperscript{1}},
 \textbf{Yun Fu\textsuperscript{1}},
\\
\\
   \textsuperscript{1}Northeastern University,
\\
}

\begin{document}
\maketitle

\input{main}

\bibliography{custom}

\newpage
\appendix

% \section{Appendix}
\label{sec:appendix}
\input{latex/appendix}

\end{document}

%% file: main.tex
\begin{abstract}
Causal reasoning in natural language requires identifying relevant variables, understanding their interactions, and reasoning about effects and interventions, often under noisy or ambiguous conditions. While large language models (LLMs) exhibit strong general reasoning abilities, they struggle to disentangle correlation from causation, particularly when observations are partially incorrect or irrelevant information is present. In this work, we introduce \textit{NoisyCausal}, a new benchmark designed to evaluate causal reasoning under structured noise. Each instance is generated from a ground-truth causal graph and contextualized with a natural language scenario by injecting controllable forms of noise, such as irrelevant distractors, value perturbations, confounding, and partial observability. Moreover, we propose a modular reasoning framework that combines LLMs with explicit causal structure to address these challenges. Our method prompts the LLM to extract variables, construct a causal graph from context, and then reformulates the reasoning task as a structured prompt grounded in this graph. Rather than relying on statistical patterns alone, the LLM is guided by symbolic structure, enabling more interpretable and robust inference. Experimental results show that our method significantly outperforms standard prompting and reasoning baselines on \textit{NoisyCausal}. Furthermore, it generalizes well to external benchmarks such as Cladder without task-specific tuning. Our findings highlight the importance of combining causal abstractions with language-driven reasoning to achieve faithful and robust causal understanding in LLMs.
\end{abstract}

\section{Introduction}
Causal reasoning is a core component of human cognition and a critical capability for intelligent systems. It enables agents to move beyond surface-level correlations to identify underlying mechanisms, support counterfactual thinking, and generalize across domains. Despite recent advances in language modeling, current systems still struggle with structured cause--effect reasoning, particularly under uncertainty or noisy observations.

Existing benchmarks \cite{bondarenko2022causalqa, jin2023cladder} primarily focus on clean, abstract scenarios where causal structure is simple or implicitly assumed. Consequently, they fail to reflect real-world conditions, where observations are often noisy, incomplete, or confounded by irrelevant information. Moreover, many recent approaches \cite{lasheras2025interventional, luo2025causal} treat causal reasoning as a generic prediction problem without explicitly modeling underlying dependencies, limiting interpretability and robustness to spurious correlations.

\begin{figure}[t]
    \centering
    \includegraphics[width=\linewidth]{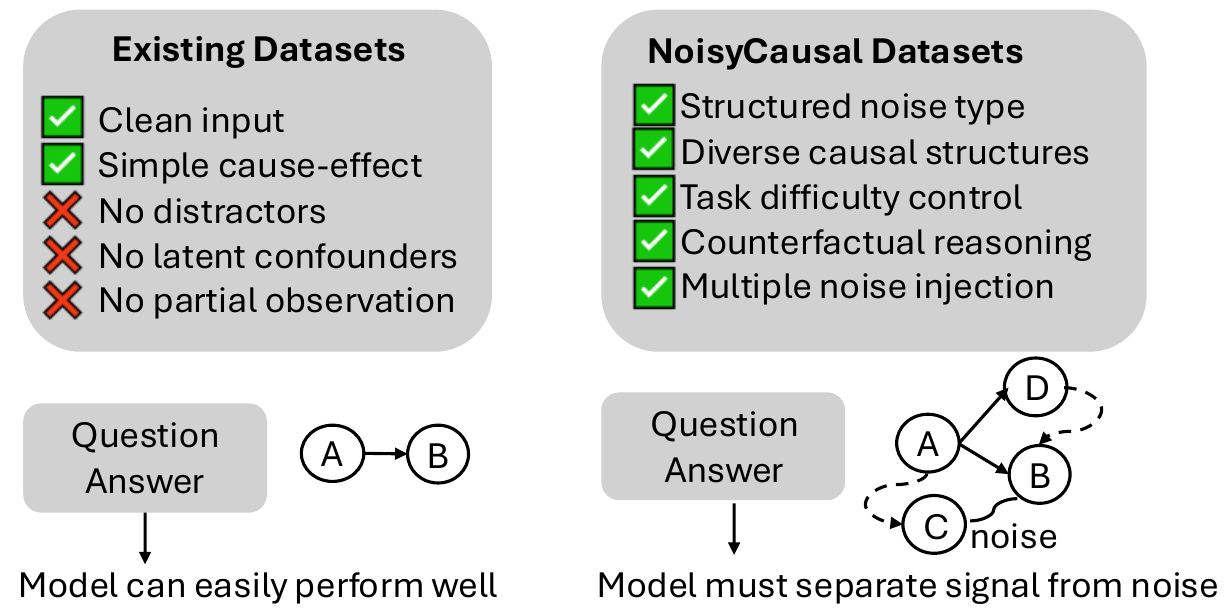}
    \caption{Existing causal reasoning datasets typically rely on clean, idealized scenarios, allowing large language models (LLMs) to succeed via pattern matching or shallow heuristics. In contrast, \textit{NoisyCausal} introduces structured noise—such as value perturbations, irrelevant variables, and latent confounders—to create more realistic and challenging reasoning environments.}
    \label{fig:iter}
    % \vspace{-20pt}
\end{figure}

To address these limitations, we introduce \textit{NoisyCausal}, a benchmark designed to evaluate causal reasoning under realistic, structured noise. Each instance is generated from an explicit causal graph and grounded in a natural language scenario (e.g., medical, mechanical, or social domains). Variable values are sampled from a structural causal model (SCM) to ensure causal consistency, after which controlled perturbations—such as irrelevant variable injection, value flipping, partial observability, and latent confounding—are applied. This design enables fine-grained evaluation of robustness, reasoning fidelity, and generalization.

Beyond the benchmark, we propose a modular causal reasoning framework that leverages large language models (LLMs) in a structured and interpretable manner. Instead of end-to-end black-box inference or symbolic propagation, our approach decomposes reasoning into variable extraction, causal graph construction, and graph-guided inference. The LLM reasons over observed variables together with a textualized causal graph, retaining linguistic flexibility while being guided by explicit causal structure. The graph serves as an interpretable intermediate that focuses attention on plausible reasoning paths.

This design offers several advantages: it improves interpretability through modular decomposition, supports reasoning under noisy or incomplete observations, and enables counterfactual and interventional analysis via simple modifications to graph structure or variable assignments. Experiments on \textit{NoisyCausal} show that integrating causal structure with LLM reasoning consistently outperforms standard prompting baselines, indicating that graph-grounded prompting is a promising direction for robust causal reasoning.

\noindent\textbf{In summary, our contributions are:}
\begin{itemize}[leftmargin=*, align=left,nosep]
% \vspace{-10pt}
\item We introduce \textit{NoisyCausal}, a benchmark for evaluating causal reasoning under structured noise, with explicit causal graphs, natural language scenarios, and controllable perturbations;
% \vspace{-10pt}
\item We propose a modular graph-guided framework that structures LLM reasoning using causal abstractions while preserving linguistic flexibility;
% \vspace{-25pt}
\item We demonstrate improved performance and robustness over standard LLM prompting approaches across diverse causal reasoning settings.
\end{itemize}

\begin{figure*}[t]
    \centering
    \includegraphics[width=0.9\linewidth]{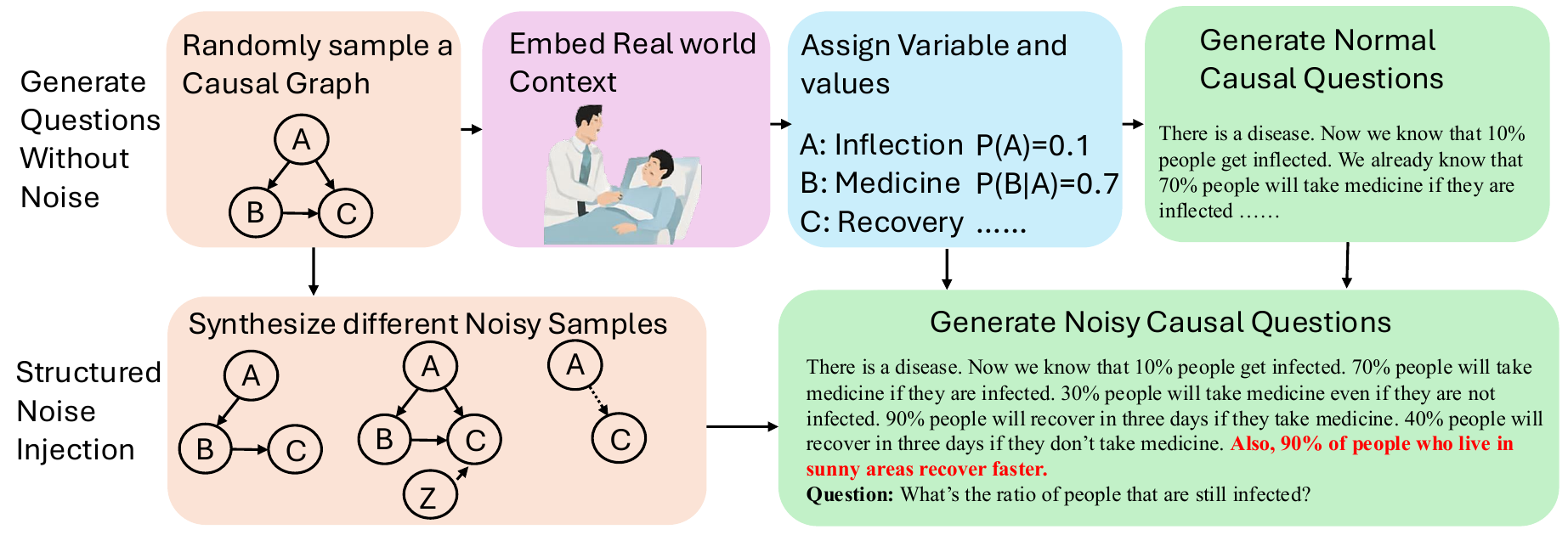}
    \caption{Overview of the \textit{NoisyCausal} dataset construction pipeline. The process begins by randomly sampling a causal graph and embedding it into a realistic real-world scenario (e.g., modeling infection, medicine intake, and recovery). Each variable is assigned a semantic role and a probabilistic function. Clean, noise-free natural language questions are then generated from this grounded causal model. To simulate real-world uncertainty, we inject structured noise—such as irrelevant variables, value perturbations, latent confounders, and missing information—into both the graph and variable observations. This leads to the synthesis of multiple noisy instances, which are then used to formulate diverse, linguistically fluent reasoning questions. The resulting dataset enables rigorous evaluation of causal reasoning robustness in large language models across varying noise conditions.}
    \label{architecture}
    \vspace{-10pt}
\end{figure*}
\section{Related Work}

\paragraph{Causal Reasoning Benchmarks.}
Evaluating models on causal reasoning has been an active research area across NLP, vision, and knowledge representation. Early work such as COPA \cite{roemmele2011choice} focuses on binary selection between plausible causes and effects, emphasizing shallow commonsense plausibility. Other datasets like Event2Mind \cite{rashkin2018event2mind} and ATOMIC \cite{sap2019atomic} extend this direction by modeling causal commonsense in social and narrative contexts. However, these benchmarks generally rely on implicit structure, and models are not required to explicitly recover or reason over a formal causal graph.

More recent datasets have attempted to bring structural causal modeling into the evaluation loop. CausalQA \cite{bondarenko2022causalqa} introduces causal reasoning questions over tabular datasets, involving do-calculus and interventions. Cladder \cite{jin2023cladder} provides an algorithm that can automatically generate causal reasoning questions. These efforts provide valuable insights but are either limited in domain diversity, lack textual grounding, or assume complete observability and clean environments. Most notably, they do not explicitly test how models handle structured noise or spurious variables that may mislead correlation-based heuristics.

\paragraph{Reasoning with Large Language Models.}
Large language models (LLMs) such as GPT-3 \cite{brown2020language}, PaLM \cite{chowdhery2023palm}, and GPT-4 \cite{achiam2023gpt} have demonstrated remarkable emergent capabilities in reasoning, planning, and few-shot generalization. Techniques such as chain-of-thought prompting \cite{wei2022chain}, tree-of-thought \cite{yao2023tree}, and ReAct \cite{yao2023react} have improved multi-step reasoning through intermediate generation. LLMs have also been used to assist with formal logic problems \cite{pan2023logic}, math \cite{zhou2023solving}, and tool use \cite{yuan2024easytool}. However, causal reasoning remains a uniquely difficult challenge for LLMs. Recent studies \cite{jiang2023large, cheng2024interactive} show that LLMs frequently confuse correlation with causation and fail to reason accurately about interventions or counterfactuals, especially in noisy or ambiguous environments. In many cases, models default to learned statistical associations from pretraining corpora, rather than identifying mechanistic dependencies. For example, they may predict that "people who cough take medicine" without understanding that coughing is an effect of illness and not a cause of recovery. These limitations highlight the need for structural guidance in the reasoning process.

\paragraph{Causal Graph Modeling and Structured Inference.}
Explicit causal modeling has a long history in statistics and AI, with foundational work such as Pearl's structural causal models \cite{pearl2009causality} and the PC/FCI family of causal discovery algorithms \cite{malinsky2018causal}. In machine learning, causal graphs have been integrated into representation learning \cite{scholkopf2021toward}, counterfactual simulation \cite{zuo2022counterfactual}, and reinforcement learning \cite{dasgupta2019causal}. Many recent approaches use graph neural networks (GNNs) to perform inference over learned or given causal graphs, such as CausalGNN \cite{wang2022causalgnn} or structure-aware transformers \cite{chen2022structure}.

\section{Dataset Construction – The \textit{NoisyCausal} Benchmark}

To rigorously evaluate causal reasoning capabilities in language models, we construct \textit{NoisyCausal}, a synthetic benchmark that couples symbolic causal structure with realistic natural language prompts and structured noise. The dataset generation pipeline is illustrated in Figure~\ref{architecture}. It consists of five sequential stages: (1) causal graph sampling, (2) semantic grounding, (3) structural causal model (SCM) sampling, (4) structured noise injection, (5) natural language question assembly.

\textbf{Causal Graph Sampling} We begin by generating a directed acyclic graph (DAG) $G = (V, E)$, where each node $v_i \in V$ represents an abstract variable and each edge $(v_i \rightarrow v_j) \in E$ denotes a direct causal influence. To encourage structural diversity, we vary graph size between 3 and 7 nodes and sample from multiple topological motifs such as chains, forks, colliders, and graphs with multiple converging parents. We use topological sorting to ensure acyclicity and discard any cyclic or disconnected graphs. Each generated graph forms the backbone of an eventual reasoning instance.

\textbf{Semantic Grounding} Once a graph is sampled, we assign each node a real-world semantic label based on predefined domains such as medicine, education, and economics. For example, a path like $A \rightarrow B \rightarrow C$ may be interpreted as ``infection causes medication intake, which affects recovery.'' These labels are drawn from a curated vocabulary and adjusted to ensure internal coherence within the scenario. In addition, we assign metadata to each variable, including its type (binary, categorical, continuous), observability (observable or latent), and role (e.g., symptom, cause, mediator, or outcome). The whole graph is then embedded into a naturalistic background story, serving as context for downstream reasoning questions.

\textbf{SCM Sampling} Given a semantically grounded causal graph, we generate a consistent assignment of variable values by sampling from a structural causal model (SCM). Each node is associated with a structural equation $f_i$ that determines its value based on its parents. The form of $f_i$ is sampled from a class of logic-based or probabilistic rules: binary nodes may use logical conjunction/disjunction, categorical nodes may follow lookup tables with probabilistic outputs, and continuous nodes are defined using additive or threshold-based functions. Sampling proceeds in topological order to respect causality. For example, given a scenario where $A$ represents infection, $B$ represents taking medicine, and $C$ represents recovery, we may specify: $P(A = 1) = 0.1$ (10\% infection rate), $P(B = 1 | A = 1) = 0.7$, $P(B = 1 | A = 0) = 0.3$ (higher medicine uptake among infected), $P(C = 1 | A = 1) = 0.4$, $P(C = 1 | B = 1) = 0.9$ (recovery probability conditioned on infection or treatment).
Such probabilistic dependencies instantiate clean observational traces that reflect the underlying causal mechanisms.

\textbf{Structured Noise Injection} To model real-world imperfections, we introduce structured noise types that distort variable values, the causal relationships, or the question context. Here, we organize the noise types from local variable-level perturbations to higher-level structural and linguistic disturbances. These include: \textbf{Value Perturbation (VP):} Randomly alter variable assignments. For example, flipping the value of $C$ (recovery) from 1 to 0 even if the structural model predicts recovery; \textbf{Irrelevant Variable Injection (IV):} Introduce non-causal variables (e.g., "drinks tea") that correlate spuriously with outcomes like recovery, misleading LLMs; \textbf{Partial Masking (PM):} Hide values of selected observed variables (e.g., $B$), simulating missing information scenarios that force the model to reason under uncertainty; \textbf{Causal Swap (CS):} Swap values of causally linked nodes (e.g., $B$ and $C$), which breaks conditional logic (e.g., makes it appear that recovery causes medicine intake); \textbf{Latent Confounders (CI):} Simulate hidden variables influencing multiple nodes (e.g., an unobserved variable $Z$ influencing both $A$ and $C$), introducing dependencies not shown in the observed graph; \textbf{Question Perturbation (QP):} Modify or corrupt the question such that its assumptions contradict the proper SCM (e.g., asking about outcomes under counterfactual settings).

Each noise type is applied under a controlled probability distribution and can be composed with others to simulate increasingly challenging inference conditions. This allows for fine-grained control over task difficulty and supports comprehensive robustness evaluation. Clean and noisy variants of each sample are stored in parallel to enable contrastive evaluation and supervision.
\begin{figure}
    \centering
    \includegraphics[width=\linewidth]{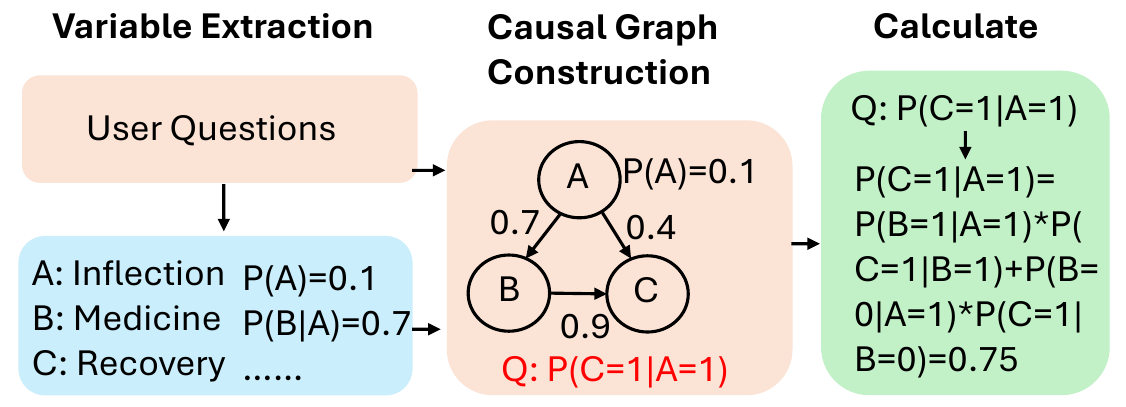}
    \caption{ Overview of the graph-guided causal reasoning framework. The process begins by extracting task-relevant variables and constructing a question-specific causal graph via LLM prompting. This graph encodes directional relationships among variables (e.g., ``infection'' causes ``medicine'', which causes ``recovery''). Observed variable values and graph structure are then converted into a structured natural language prompt, which is passed to an LLM for final inference. 
}
    \label{fig:enter-label}
    \vspace{-10pt}
\end{figure}

\textbf{Natural Language QA Assembly} Finally, we transform each data instance into a natural language prompt and a reasoning question. The generated text explicitly or implicitly reflects the assigned probabilities and variable interactions defined in Step 3 for each scenario. For instance, a prompt may state: "Only 10\% of people are infected. Among those infected, 70\% take medicine. Medicine leads to recovery in 90\% of cases. What is the likelihood that someone recovers if infected and takes medicine?" We support several task types, including interventional queries ("If the patient had not taken medicine, what would happen?"), counterfactual reasoning ("Had the infection not occurred, would recovery still be likely?"), Moreover, attributional analysis ("What caused the recovery?"). The background scenario is narrated fluently, and variable names are embedded into the text using handcrafted templates and paraphrasing engines to produce diverse and natural questions. Answers are derived based on the clean SCM and the causal graph, even in the presence of distractors or inconsistent evidence.

\section{Causal Graph-Based Reasoning Model}

\vspace{-5pt}
To enable robust causal reasoning under noisy natural language conditions, we propose a framework that integrates lightweight causal graph construction with large language models (LLMs). Instead of relying on symbolic inference or message-passing algorithms, we use causal graphs as intermediate, interpretable structures that organize relevant variables and their directional relationships. The LLM then performs reasoning based on this structured representation.

The process begins with variable extraction, where the LLM identifies key entities (e.g., “fever”, “medicine”) and latent factors (e.g., “infection”) from the background context. Next, we construct a task-specific causal graph using LLM-based prompting: for each variable pair, the model predicts causal directionality, forming a directed acyclic graph. Low-confidence or inconsistent edges are post-processed or deferred.

Once the graph is built, we compile a natural language prompt containing the background, observed variable values, and a textual description of the graph. The LLM uses this structured input to answer the reasoning question, producing answers that reflect both surface context and the underlying causal structure.

This architecture enhances explainability, supports counterfactual queries via prompt edits, and improves robustness under noise. By separating structure extraction from inference, our method encourages causal reasoning over correlation, bridging symbolic structure and LLM flexibility.

\section{Experiments}

\begin{table*}[!htb]
\centering
\caption{
Accuracy (\%) of different methods across datasets and noise types. The left block shows performance on our \textit{NoisyCausal} benchmark under different noise types (VP: Value Perturbation, IV: Irrelevant Variable, etc.); the right block shows generalization to Cladder without task-specific tuning.
}
\label{tab:combined-results}
\resizebox{0.98\linewidth}{!}{
\begin{tabular}{lcccccccc}
\toprule
\multirow{2}{*}{\textbf{Model}} & \multicolumn{7}{c}{\textbf{NoisyCausal}} & \textbf{External Dataset} \\
& W/O Noise & VP & IV & CS & PM & CI & QP & Cladder \\
\midrule
GPT-3.5\cite{brown2020language} & 57.9 & 54.2 & 49.6 & 44.7 & 52.0 & 47.1 & 43.8 & 52.2 \\
GPT-4\cite{achiam2023gpt}   & 62.8 & 60.4 & 56.3 & 50.1 & 59.8 & 54.0 & 48.5 & 62.0 \\
LLaMa\cite{touvron2023llama}   & 52.4 & 50.1 & 45.7 & 41.3 & 48.9 & 42.6 & 40.7 & 44.0 \\
Alpaca\cite{taori2023alpaca}  & 53.6 & 51.4 & 46.6 & 42.0 & 50.3 & 43.8 & 41.2 & 44.7 \\
CoT\cite{wei2022chain}    & 65.5 & 63.5 & 58.9 & 55.4 & 60.8 & 56.2 & 53.3 & -- \\
ToT\cite{yao2023tree}     & 68.3 & 65.1 & 61.7 & 57.2 & 63.4 & 59.5 & 54.6 & -- \\
ReAct\cite{yao2023react}   & 64.5 & 64.2 & 60.4 & 56.1 & 62.5 & 57.0 & 52.7 & -- \\
Reflexion\cite{shinn2023reflexion} & 67.8 & 65.7 & 62.3 & 58.4 & 64.1 & 60.2 & 56.9 & -- \\
Causal CoT\cite{jin2023cladder} & 73.4 & 70.4 & 66.9 & 63.1 & 68.8 & 65.2 & 60.7 & 70.4 \\
\textbf{Ours (Graph-Guided)} & \textbf{80.7} & \textbf{77.3} & \textbf{74.6} & \textbf{71.8} & \textbf{76.2} & \textbf{73.5} & \textbf{69.9} & \textbf{82.3} \\
\bottomrule
\end{tabular}
}
\end{table*}

\subsection{Datasets}

Our primary evaluation dataset is \textbf{NoisyCausal}, a synthetic benchmark containing 10,617 question-answer pairs with controlled causal structures and structured noise injections. Each instance includes a natural language background, an observation set (clean or perturbed), a causal reasoning question, and the ground-truth answer derived from a symbolic SCM. We evaluate models under six primary noise types (Value Perturbation, Irrelevant Variables, Causal Swap, Partial Masking, Confounders, and Question Perturbation) and multi-noise combinations of increasing difficulty. We also report results on the \textbf{Cladder} datasets for generalization.

\subsection{Main Results}

Table~\ref{tab:combined-results} presents the performance of various models on the \textit{NoisyCausal} benchmark and the external Cladder dataset. Our graph-guided method achieves the highest accuracy across all conditions, including both clean and noisy settings. In the noise-free case, it reaches 80.7\% accuracy, outperforming GPT-4 (62.8\%) and Causal CoT (73.4\%), indicating that structural guidance improves even standard reasoning. Under structured noise types such as value perturbation (VP), irrelevant variables (IV), and latent confounders (CI), traditional LLMs and CoT-style methods show noticeable degradation, with GPT-3.5 falling to as low as 43.8\% on question perturbation (QP). In contrast, our model maintains stable performance across all perturbations (e.g., 77.3\% on VP, 74.6\% on IV, 73.5\% on CI), highlighting its robustness to uncertainty. While advanced prompting strategies like ToT, ReAct, and Reflexion offer moderate gains over vanilla prompting, they are still sensitive to causal inconsistencies such as swaps or confounders. Finally, our model generalizes well to Cladder, achieving 82.3\% accuracy without task-specific tuning, outperforming all baselines. These results demonstrate the benefit of combining causal structure with LLMs for robust and interpretable reasoning under noisy conditions.

\begin{table}[!htt]
\centering
\caption{
Ablation study results on the NoisyCausal benchmark (Accuracy \%). Each variant removes or alters a specific component of our full model to evaluate its contribution. The final block shows the performance drop as more types of noise are combined.
}
\label{tab:ablation-updated}
\resizebox{0.5\textwidth}{!}{
\begin{tabular}{lcc}
\toprule
\textbf{Ablation Setting} & \textbf{Accuracy (\%)} & \textbf{ vs. Full Model} \\
\midrule
\textbf{Full Model (Ours)}                & \textbf{80.68} & -- \\
\midrule
\multicolumn{3}{l}{\textit{Graph Structure and Variable Semantics}} \\
\quad No Graph                            & 65.32 & -15.36 \\
\quad Random Graph                        & 60.87 & -19.81 \\
\quad Shuffled Variable Names             & 63.41 & -17.27 \\
\midrule
\multicolumn{3}{l}{\textit{Prompt Design and Language Robustness}} \\
% \quad Structured Prompt                   & 79.23 & -1.45 \\
\quad Edge-Only Prompt                    & 74.34 & -6.34 \\
\quad Natural Prompt                      & 78.92 & -1.76 \\
\quad Question Variation                  & 80.12 & -0.56 \\
\midrule
\multicolumn{3}{l}{\textit{Cumulative Noise Combinations}} \\
\quad 1 Noise Type                        & 73.47 & -7.21 \\
\quad 2 Noise Types                       & 67.32 & -13.36 \\
\quad 3 Noise Types                       & 63.45 & -17.23 \\
\quad 4 Noise Types                       & 60.51 & -20.17 \\
\quad All 6 Noise Types                   & 57.98 & -22.70 \\
\bottomrule
\end{tabular}
}
\end{table}

\subsection{Ablation Studies}

To better understand the effectiveness and robustness of our proposed framework, we conduct a comprehensive set of ablation studies. These experiments are designed to isolate and evaluate the contributions of individual components within the system, including the causal graph, prompt design, and resilience under increasingly noisy conditions. As shown in Table \ref{tab:ablation-updated}, our analysis is structured into three categories.

\paragraph{Graph Structure and Variable Semantics.}
We first examine how the presence and correctness of causal structure influence reasoning. Removing the causal graph entirely (\textit{No Graph}) results in significant performance degradation, indicating that the model falls back on shallow statistical heuristics. Replacing the ground-truth graph with a structurally equivalent but randomly rewired version (\textit{Random Graph}) leads to even lower accuracy, confirming that not only the existence of structure but its correctness is crucial. We also test the impact of variable naming by shuffling variable names across the graph (\textit{Shuffled Variable Names}), which disrupts semantic alignment with the question and causes further degradation. These results suggest that structural topology and linguistic consistency contribute substantially to model performance.

\paragraph{Prompt Design and Language Robustness.}
We evaluate the effect of different ways of presenting the graph on the model. Switching from natural language descriptions to a structured prompt (full model) format improves robustness slightly. Providing the graph as an edge list alone (\textit{Edge-Only Prompt}) hurts performance, likely due to the lack of reasoning context. We also test \textit{Question Variation} by paraphrasing the same prompt in different syntactic forms. The model maintains stable performance, suggesting it generalizes well to surface-level linguistic changes.

\paragraph{Cumulative Noise Combinations.}
To assess the model's robustness under compounding uncertainty, we introduce a new ablation protocol where noise types are randomly sampled and composed. Instead of isolating individual noise categories (e.g., Value Perturbation or Confounder Injection), we incrementally combine multiple types to simulate more realistic and challenging environments. We observe a graceful decline in performance as the number of noise types increases: from 73.47\% accuracy with one noise type to 57.98\% when all six types are combined.

\paragraph{Summary}
Together, these ablation studies highlight several insights: (1) the presence and correctness of the causal graph structure play a central role in reasoning quality, (2) LLMs benefit from natural language contextualization of structure, not just formal edge lists, (3) robustness varies significantly across different types of noise, with distractors and confounders posing the most serious challenges, and (4) variable and graph extraction modules must be accurate and well-calibrated, as small mistakes can propagate downstream. These findings further support the design philosophy of using causal structure to constrain and guide LLM-based reasoning under uncertainty.

\subsection{Graph Perturbation Sensitivity}

In ablation studies, we noticed that the correctness of the causal graph structure plays a central role in reasoning quality. Hence, we want to explore the graph perturbation sensitivity further. Here, we perform a controlled structural perturbation analysis. Starting from the ground-truth graph, we introduce common types of errors: edge deletion (ED), where key connections are removed; false edge injection (FE), where spurious edges are added; and edge direction reversal (DR), which flips the cause-and-effect direction of existing edges. As shown in Figure \ref{fig:graph-perturbation}, we evaluate three types of structural noise: Edge Deletion, False Edge Injection, and Direction Reversal, each applied with increasing numbers of errors (1–4). we find that while the model is relatively robust to minor edge deletion, it is significantly more sensitive to misleading or reversed causal links. In particular, direction reversal leads to the steepest performance drop, indicating that incorrect directional flow severely disrupts reasoning chains. These results emphasize the importance of accurate graph construction and motivate future research in validating or refining LLM-generated causal graphs before inference.

\subsection{Structure Discovery Reliability}

A key component of our framework is the ability to automatically construct causal graphs from natural language using a prompted large language model (LLM). While our downstream performance suggests that these structures are often useful, we seek to evaluate the accuracy and reliability of LLM-extracted graphs directly. To this end, we select 1000 NoisyCausal examples with known ground-truth graphs and prompt the LLM to extract relevant variables and causal edges. We compare the predicted graphs to the ground truth using edge-level Precision, Recall, and F1 score.

We evaluate performance across three axes: (1) different noise levels (clean vs. perturbed vs. masked/confounded), (2) prompt styles (structured vs. natural language), and (3) model types (GPT-4 vs. LLaMA). Results in Table~\ref{tab:structure-discovery} show that GPT-4 achieves high accuracy in clean settings (F1 > 85) and remains robust under moderate noise, while smaller models like LLaMA degrade significantly. Structured prompts yield better extraction quality than open-ended instructions. These findings suggest that LLMs can recover causal structure with reasonable fidelity, but remain sensitive to input noise and prompt formulation—highlighting the need for better prompt calibration, graph validation, or hybrid symbolic support.
\begin{table}[t]
\centering
\caption{Structure discovery performance across models, noise levels, and prompt designs. Reported: Edge-level Precision / Recall / F1 (\%).}
\label{tab:structure-discovery}
\resizebox{0.47\textwidth}{!}{
\begin{tabular}{lccc}
\toprule
\textbf{Condition} & \textbf{Precision} & \textbf{Recall} & \textbf{F1 Score} \\
\midrule
GPT-4 (clean, structured prompt) & 87.1 & 83.5 & 85.2 \\
GPT-4 (value perturbed) & 81.3 & 78.9 & 80.1 \\
GPT-4 (confounding + masking) & 76.2 & 72.1 & 74.1 \\
GPT-4 (clean, natural prompt) & 82.5 & 77.4 & 79.8 \\
\midrule
LLaMA-2 (clean, structured prompt) & 65.4 & 60.7 & 62.9 \\
LLaMA-2 (value perturbed) & 60.4 & 57.6 & 58.9 \\
LLaMA-2 (confounding + masking) & 52.3 & 49.6 & 50.9 \\
\bottomrule
\end{tabular}
}
\vspace{-5pt}
\end{table}

\subsection{Error Typology Analysis}
To gain deeper insights into the types of failures exhibited by our model, we conduct a qualitative and quantitative analysis of model errors across the NoisyCausal benchmark. Specifically, we manually inspect a representative subset of incorrect predictions and categorize them into distinct error types based on their underlying cause. This analysis allows us to better understand how different forms of reasoning breakdown occur and how they relate to specific challenges introduced by structured noise. We define the following error categories: \textbf{E1. Variable Identification Failure}: The model fails to recognize or track a key variable mentioned in the prompt (e.g., overlooks a latent confounder or misidentifies the target outcome); \textbf{E2. Causal Path Misinterpretation}: The model identifies variables but incorrectly infers the causal direction or reasoning chain (e.g., infers recovery → medicine instead of medicine → recovery); \textbf{E3. Numerical Reasoning Error}: The model identifies the structure correctly but fails to compute or compare probabilities accurately, often due to surface-level heuristics; \textbf{E4. Distractor Overfitting}: The model is misled by irrelevant or spurious variables introduced as structured noise, such as "people who drink tea recover faster"; \textbf{E5. Question Misalignment}: The model misunderstands the logical intent of the question, especially in counterfactual or interventional settings.

\begin{figure}[t]
    \centering
    \begin{minipage}[t]{0.48\linewidth}
        \centering
        \includegraphics[width=\linewidth]{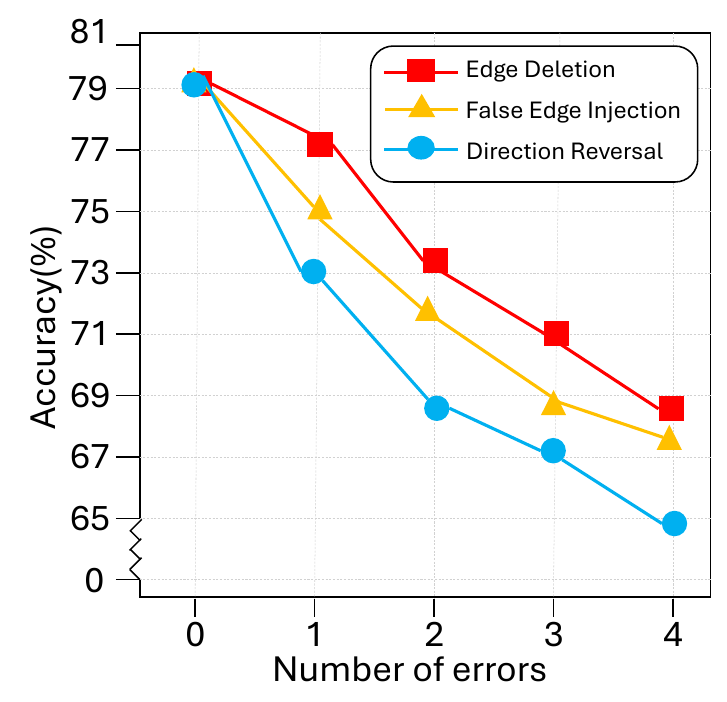}
        \caption{Impact of causal graph perturbations on reasoning accuracy. }
        \label{fig:graph-perturbation}
    \end{minipage}
    \hfill
    \begin{minipage}[t]{0.48\linewidth}
        \centering
        \vspace*{-7\baselineskip}
        \includegraphics[width=0.8\linewidth]{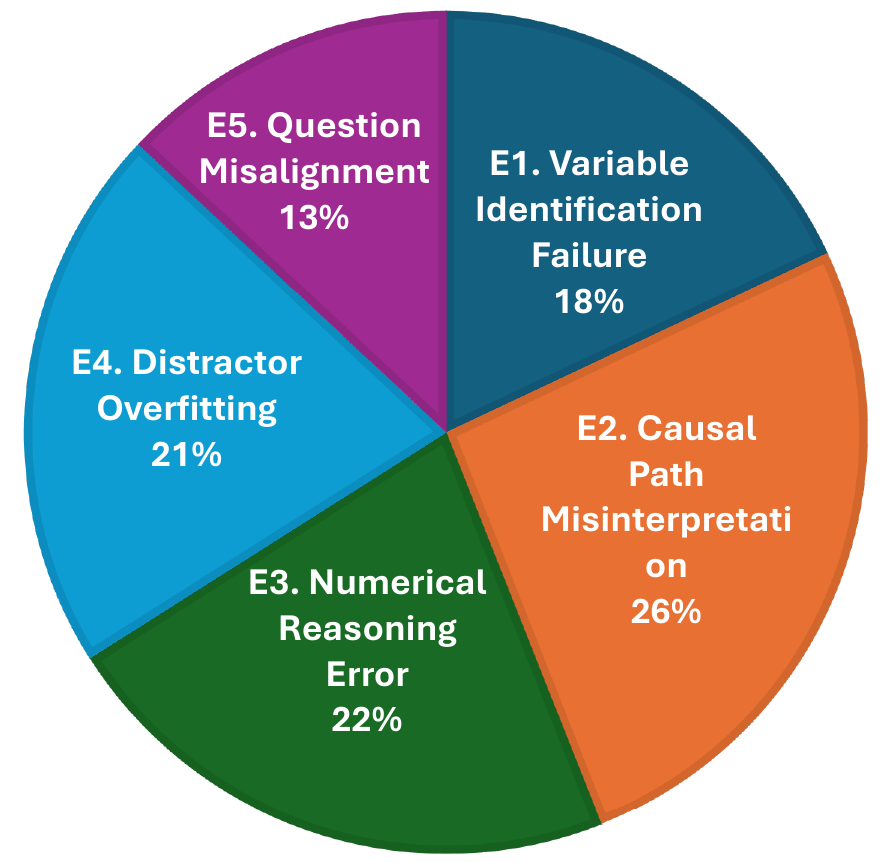}
        \caption{Distribution of error types across a sample of 100 incorrect predictions.}
        \label{fig:distribution}
    \end{minipage}

\vspace{-10pt}
\end{figure}

We randomly sample 100 error cases from the validation set across all six noise types and assign each error to one of the five categories. The results are summarized in Figure \ref{fig:distribution}

The most common failure mode involves incorrect causal path interpretation (E2), especially in the presence of swapped edges or latent confounding. This highlights the need for stronger graph validation or constraint mechanisms. A substantial portion of errors are also due to distractor overfitting (E4), suggesting that LLMs still struggle to distinguish spurious associations even when guided by structure. Interestingly, numerical reasoning errors (E3) are not limited to statistical estimation, but also occur when conflicting cues (e.g., base rates vs conditional probabilities) are present.

These findings underscore the multifaceted nature of causal reasoning errors and suggest that future improvements should not only target graph construction accuracy, but also incorporate better reasoning supervision, distractor suppression, and probabilistic calibration.

\subsection{Belief-Inconsistent Perturbation}

We design a new experiment to evaluate model robustness when exposed to misleading but structurally irrelevant contextual beliefs. Specifically, we introduce \textbf{Belief-Inconsistent Perturbation (BIP)}, where the prompt includes incorrect or commonly held misconceptions that contradict the underlying causal graph, while the actual variable relationships and numerical values remain unchanged. For example, the scenario may state “Many people believe that taking medicine increases the chance of infection,” even though the true causal direction is \texttt{Infection $\rightarrow$ Medicine}. We inject such statements into 1000 clean NoisyCausal instances and compare performance with the original versions. As shown in Table~\ref{tab:bip}, models relying on surface-level reasoning (e.g., GPT-3.5, GPT-4) show notable accuracy drops (up to 6\%), suggesting vulnerability to belief-level cues. In contrast, our graph-guided model degrades minimally, demonstrating that structural grounding improves resistance to contextual bias.

\begin{table}[t]
\centering
\caption{\small Performance under Belief-Inconsistent Perturbation (BIP). Accuracy (\%) on original and belief-perturbed questions.}
\label{tab:bip}
\renewcommand{\arraystretch}{0.6} % reduce row height
\resizebox{0.9\linewidth}{!}{
\begin{tabular}{lcc}
\toprule
\textbf{\small Model} & \textbf{\small Original} & \textbf{\small + BIP} \\
\midrule
\small GPT-3.5 & \small 57.9 & \small 51.2 \\
\small GPT-4 & \small 62.8 & \small 56.7 \\
\small Causal CoT & \small 73.4 & \small 69.1 \\
\small Ours (Graph-Guided) & \textbf{\small 80.7} & \textbf{\small 79.5} \\
\bottomrule
\end{tabular}
}
\vspace{-10pt}
\end{table}

\subsection{Transferability to Real-World Causal Reasoning}

A common concern with synthetic benchmarks is whether improvements observed under controlled settings can transfer to real, human-authored causal reasoning tasks. We emphasize that \textit{NoisyCausal} is not intended to replace real-world datasets, but rather to function as a diagnostic benchmark that isolates specific failure modes—such as distractor sensitivity, confounding, and partial observability—that are difficult to disentangle in natural corpora.
To evaluate whether exposure to structured causal noise yields transferable benefits, we conduct a small-scale transfer experiment on \textbf{CausalQA}, a human-authored benchmark involving natural language causal questions. We fine-tune a \textbf{LLaMA-7B} model using LoRA on \textit{NoisyCausal}, and then directly evaluate the resulting model on CausalQA without any task-specific tuning or adaptation.

We compare a zero-shot LLaMA-7B baseline evaluated directly on CausalQA with a variant that is LoRA fine-tuned on \textit{NoisyCausal} prior to evaluation. As shown in Table~\ref{tab:transfer-causalqa}, LoRA fine-tuning on \textit{NoisyCausal} improves accuracy on causal questions in CausalQA from 44.2\% to 49.3\%, yielding a gain of 5.1 percentage points. Notably, the model is never exposed to CausalQA during training, indicating that the observed improvement arises from transferable causal reasoning skills rather than task memorization.

\begin{table}[t]
\centering
\caption{Transfer results on CausalQA (causal questions only).}
\label{tab:transfer-causalqa}
\renewcommand{\arraystretch}{1.1}
\resizebox{0.95\linewidth}{!}{
\begin{tabular}{lcc}
\toprule
\textbf{Model} & \textbf{Training Data} & \textbf{Accuracy (\%)} \\
\midrule
LLaMA-7B (baseline) & -- & 44.2 \\
+LoRA (NoisyCausal) & NoisyCausal & 49.3 \; (+5.1) \\
\bottomrule
\end{tabular}
}
\vspace{-10pt}
\end{table}

These results suggest that although \textit{NoisyCausal} is synthetic by construction, the structured noise it introduces captures aspects of real-world causal ambiguity that are beneficial for generalization. In this sense, \textit{NoisyCausal} serves as an effective stress-test and auxiliary training signal for improving robustness in causal reasoning, rather than a task-specific synthetic shortcut.

\section{Conclusion}

We introduced NoisyCausal, a benchmark for evaluating LLMs' causal reasoning under structured noise. Each instance is grounded in a causal graph and perturbed with distractors, confounders, or inconsistent information to assess robustness and reasoning fidelity. To address this challenge, we proposed a modular graph-guided framework that extracts variables and constructs causal graphs from natural language, reformulating the task into structured prompts for more interpretable and noise-resilient inference. Experiments show our method consistently outperforms standard prompting and generalizes well to external datasets like Cladder. Ablation studies highlight the importance of accurate structure and prompt design. Looking forward, NoisyCausal can support future research in causally aware LLMs, including extensions to real-world and multimodal tasks, and improvements in efficiency via lightweight reasoning agents. Our findings demonstrate the value of combining symbolic structure with language-based reasoning for robust and faithful AI.

\section{Limitations}

While our framework shows strong performance under noisy conditions, several limitations remain. First, it relies on LLM-based variable extraction and graph construction, which may introduce cascading errors. Future work could incorporate confidence estimation or correction mechanisms to improve stability. Second, the current modular design lacks end-to-end optimization. Joint training of variable extractors and graph builders, or lightweight finetuning strategies, may enhance robustness in low-resource settings. Third, although \textit{NoisyCausal} includes diverse synthetic noise, real-world generalization remains untested. Future benchmarks with real or multimodal data (e.g., tables, images) are needed to evaluate broader applicability. Finally, our method assumes a recoverable structure from the input. Open-domain or discourse-level tasks may require stronger integration with causal discovery or contrastive learning to infer latent structure.

\noindent\textbf{AI Disclosure} We used ChatGPT solely for grammar correction and language polishing. All research content, literature analysis, and writing were conducted independently by the authors.

%% file: latex/appendix.tex
\section{Dataset Composition and Annotation.} 
Each \textit{NoisyCausal} instance includes: (1) a natural language background and reasoning question, (2) a corresponding causal graph $G$ (optionally provided to models), (3) both clean and noisy variable assignments, (4) the correct answer computed from the SCM, and (5) metadata describing the applied noise types and severity. These components allow \textit{NoisyCausal} to support robust benchmarking of causal reasoning across structured noise conditions, enabling diagnostic insights into model behavior under uncertainty.

\section{Example of different noise injection}

\begin{table*}[!t]
\centering
\scriptsize
\renewcommand{\arraystretch}{1.2}
\begin{tabular}{p{2.5cm} p{6.5cm} p{2.5cm} p{2.5cm}}
\toprule
\textbf{Noise Type} & \textbf{Modified Question (red = perturbed)} & \textbf{What is Perturbed} & \textbf{Potential Impact} \\
\midrule

VP: Value Perturbation &
\begin{minipage}[t]{\linewidth}
\scriptsize\raggedright
There is a disease.\\
Now we know that 10\% people get infected.\\
70\% people will take medicine if they are infected.\\
30\% people will take medicine even if they are not infected.\\
\textcolor{red}{80\%} people will recover in three days if they take medicine.\\
40\% people will recover in three days if they don’t take medicine.\\
Question: What’s the ratio of people that are still infected?
\end{minipage}
&
Numerical value of \( P(C \mid B) \) changed from 90\% to 80\%
&
Underestimates true effect of treatment, weakening causal inference. \\

\midrule

IV: Irrelevant Variable Injection &
\begin{minipage}[t]{\linewidth}
\scriptsize\raggedright
There is a disease.\\
Now we know that 10\% people get infected.\\
70\% people will take medicine if they are infected.\\
30\% people will take medicine even if they are not infected.\\
90\% people will recover in three days if they take medicine.\\
40\% people will recover in three days if they don’t take medicine.\\
\textcolor{red}{Also, 90\% of people who live in sunny areas recover faster.}\\
Question: What’s the ratio of people that are still infected?
\end{minipage}
&
Injection of irrelevant but highly correlated variable
&
Model may confuse correlation with causation, leading to spurious inference. \\

\midrule

CS: Causal Swap &
\begin{minipage}[t]{\linewidth}
\scriptsize\raggedright
There is a disease.\\
Now we know that 10\% people get infected.\\
\textcolor{red}{People who recover are more likely to have taken medicine.}\\
70\% people will take medicine if they are infected.\\
30\% people will take medicine even if they are not infected.\\
90\% people will recover in three days if they take medicine.\\
40\% people will recover in three days if they don’t take medicine.\\
Question: What’s the ratio of people that are still infected?
\end{minipage}
&
Causal direction reversed (\( C \rightarrow B \) instead of \( B \rightarrow C \))
&
Misleads the model to reverse causal flow, impacting reasoning fidelity. \\

\midrule

PM: Partial Masking &
\begin{minipage}[t]{\linewidth}
\scriptsize\raggedright
There is a disease.\\
Now we know that 10\% people get infected.\\
\textcolor{red}{(Missing) 70\% people will take medicine if they are infected.}\\
30\% people will take medicine even if they are not infected.\\
90\% people will recover in three days if they take medicine.\\
40\% people will recover in three days if they don’t take medicine.\\
Question: What’s the ratio of people that are still infected?
\end{minipage}
&
Removal of conditional dependency \( P(B \mid A) \)
&
Model must reason under uncertainty or fail gracefully. \\

\midrule

CI: Confounder Injection &
\begin{minipage}[t]{\linewidth}
\scriptsize\raggedright
There is a disease.\\
Now we know that 10\% people get infected.\\
70\% people will take medicine if they are infected.\\
30\% people will take medicine even if they are not infected.\\
90\% people will recover in three days if they take medicine.\\
40\% people will recover in three days if they don’t take medicine.\\
\textcolor{red}{Also, people with strong immune systems tend to both recover more quickly and are less likely to take medicine.}\\
Question: What’s the ratio of people that are still infected?
\end{minipage}
&
Latent confounder \( Z \) added (\( Z \rightarrow B \), \( Z \rightarrow C \))
&
Introduces backdoor path, potentially biases causal reasoning. \\

\midrule

QP: Question Perturbation &
\begin{minipage}[t]{\linewidth}
\scriptsize\raggedright
There is a disease.\\
Now we know that 10\% people get infected.\\
70\% people will take medicine if they are infected.\\
30\% people will take medicine even if they are not infected.\\
90\% people will recover in three days if they take medicine.\\
40\% people will recover in three days if they don’t take medicine.\\
Question: \textcolor{red}{If people did not get infected but still took medicine, will they definitely recover?}
\end{minipage}
&
Question contradicts causal structure
&
Forces model to answer counterfactual inconsistent with given SCM. \\

\bottomrule
\end{tabular}
\caption{Illustration of six structured noise types in NoisyCausal. Modified questions include targeted perturbations (highlighted in red) that affect different components of causal reasoning, such as structure, semantics, or inference assumptions.}
\label{tab:noise-minipage}
\end{table*}

To supplement the formal definitions of structured noise in the main text, Table~\ref{tab:noise-minipage} presents concrete examples illustrating each perturbation type used in the NoisyCausal benchmark. For clarity, all examples are based on a common question template, with only one modification applied per instance. The modified elements are highlighted in red to isolate the injected noise.

Beyond showcasing what is perturbed, we also annotate which part of the causal reasoning pipeline is likely to be affected—such as conditional dependencies, causal directions, variable observability, or question alignment. These annotations help clarify how each type of noise challenges specific reasoning skills, including structure tracking, probabilistic reasoning, confounder adjustment, and counterfactual inference.

This appendix serves as a reference to understand better how different noise patterns are instantiated in our dataset and how they may induce distinct failure modes in language-based causal reasoning models.

\section{Faliure Cases With Graph Perturbation}
Table~\ref{tab:failure-cases} illustrates representative failure cases in causal reasoning arising from incorrect or incomplete causal graph structures. While the numerical calculations may appear sound in both examples, the underlying assumptions encoded in the graph are flawed—leading to misinterpretation of the causal mechanisms.

In the first case, the model incorrectly assumes that taking medicine (X) causes infection status (
Z), inverting the actual causal direction. Although the resulting infection rate appears numerically correct, this structure violates causal semantics and will fail to generalize to interventions or counterfactual queries.
In the second case, the model omits a critical confounder (S)—student ability—which influences both the treatment (tutoring, 
T) and the outcome (exam pass, E). Without adjusting for this hidden variable, the estimated treatment effect conflates correlation with causation, yielding biased results.

These examples demonstrate that correct causal reasoning depends not only on correct math, but on correctly structured causal assumptions. They further underscore the importance of evaluating both the output and the underlying causal model in language-based reasoning systems.

\section{Fragility of Graph Induction and Oracle Upper Bound}

A key concern raised by reviewers is the fragility of the graph induction step, as errors in the inferred causal structure may propagate to downstream reasoning. To better characterize this effect, we conduct a controlled perturbation study that isolates the impact of different types of structural errors on reasoning performance.

\paragraph{Experimental Setup.}
Starting from ground-truth causal graphs in \textit{NoisyCausal}, we introduce controlled, single-edge perturbations while keeping all other components unchanged. We consider three common error types: \textbf{Edge Deletion (ED)}, where one true causal edge is removed; \textbf{False Edge Injection (FE)}, where one spurious edge is added; and \textbf{Direction Reversal (DR)}, where the direction of one existing edge is flipped. We evaluate accuracy on 1000 validation instances and report results relative to an oracle setting in which the model is provided with perfectly correct causal graphs.

\paragraph{Results.}
Table~\ref{tab:graph-fragility} summarizes the results. Under the oracle graph setting, the model achieves an accuracy of 85.0\%, representing an upper bound when causal structure is perfectly specified. Introducing a single edge deletion results in a modest performance drop of 3.3 points, indicating that the model can tolerate minor missing structural information. In contrast, false edge injection and direction reversal lead to substantially larger degradations (–8.2 and –11.8 points, respectively), suggesting that misleading or incorrect causal directions are significantly more harmful than incomplete graphs.

\begin{table}[t]
\centering
\caption{Impact of controlled graph perturbations on reasoning accuracy (1000 validation instances).}
\label{tab:graph-fragility}
\renewcommand{\arraystretch}{0.9}
\resizebox{0.9\linewidth}{!}{
\begin{tabular}{lcc}
\toprule
\textbf{Graph Condition} & \textbf{Accuracy (\%)} & $\boldsymbol{\Delta}$ \textbf{vs.\ Oracle} \\
\midrule
Oracle Graph (no error) & 85.0 & -- \\
Edge Deletion (ED) & 81.7 & --3.3 \\
False Edge Injection (FE) & 76.8 & --8.2 \\
Direction Reversal (DR) & 73.2 & --11.8 \\
\bottomrule
\end{tabular}
}
\end{table}
\begin{table*}[!h]
\centering
\caption{Failure Cases in Causal Reasoning Due to Incorrect Graph Structures}
\begin{tabularx}{\textwidth}{p{2cm}Xp{1.5cm}p{1.5cm}}
\hline
\textbf{Error Type} & \textbf{Original Question and Reasoning Error} & \textbf{Incorrect Causal Graph} & \textbf{Correct Causal Graph} \\
\hline

Graph Perturbation (X $\rightarrow$ Z) 
& 
10\% of people are infected. 70\% of infected take medicine; 30\% of uninfected also take medicine. 90\% recover in 3 days if treated; 40\% recover otherwise. \newline
\textbf{Question:} What is the infection rate? \newline
\textbf{Incorrect reasoning:} \newline
1. $P(X=1) = 0.7 \cdot 0.1 + 0.3 \cdot 0.9 = 0.34$ \newline
2. $P(Z=1|X=1) = \frac{0.7 \cdot 0.1}{0.34} \approx 0.2059$ \newline
3. $P(Z=1|X=0) = \frac{0.3 \cdot 0.1}{0.66} \approx 0.0455$ \newline
4. Final: $P(Z=1) \approx 0.2059 \cdot 0.34 + 0.0455 \cdot 0.66 \approx 0.10$ \newline
\textbf{Flaw:} Wrong causal direction (X causes Z), leads to incorrect inversion logic. \newline
\textbf{Correct Answer} : 0.025
& 
X $\rightarrow$ Z \newline
Z $\rightarrow$ Y \newline
X $\rightarrow$ Y 
& 
Z $\rightarrow$ X \newline
Z $\rightarrow$ Y \newline
X $\rightarrow$ Y \\
\hline

Graph Perturbation(Delete Edge S $\rightarrow$ E)
& 
25\% of students received tutoring. 80\% of them passed; 50\% passed without it. 40\% of students have strong ability and always pass, regardless of tutoring. \newline
\textbf{Question:} What is the pass rate? \newline
\textbf{Incorrect reasoning:} \newline
Enumerate combinations:\newline
S=1,T=1: $0.4 \cdot 0.25 \cdot 0.8 = 0.08$ \newline
S=1,T=0: $0.4 \cdot 0.75 \cdot 1 = 0.30$ \newline
S=0,T=1: $0.6 \cdot 0.25 \cdot 0.8 = 0.12$ \newline
S=0,T=0: $0.6 \cdot 0.75 \cdot 0.5 = 0.225$ \newline
Total: $P(E=1) = 0.08 + 0.30 + 0.12 + 0.225 = 0.725$ \newline
\textbf{Flaw:} S is a confounder for both T and E, and not controlling for it leads to misleading observational estimates.\newline
\textbf{Correct Answer: }0.745
& 
T $\rightarrow$ E
& 
S $\rightarrow$ E \newline
T $\rightarrow$ E \\
\hline

\end{tabularx}
\label{tab:failure-cases}
\end{table*}

\newpage
\subsection{Prompt Templates}

\paragraph{Part A: Dataset Construction}
\begin{quote}\ttfamily\small
Step 1: Causal Graph Sampling \\
You are a graph generator. \\
Given the desired number of nodes (3--7) and allowed motifs (chain, fork, collider),
you must generate a connected, acyclic directed graph (DAG) with nodes and edges, and provide a valid topological order. \\
\\
Step 2: Semantic Grounding \\
You are a semantic mapper. \\
Given the sampled graph, you must assign each node a realistic semantic role, specify type (binary/categorical/continuous), observability (observed/latent), and causal role (cause/mediator/outcome), and describe the scenario in natural language. \\
\\
Step 3: SCM Sampling \\
You are a causal model designer. \\
Given the graph and semantic mapping, you must define conditional probability distributions for each variable based on its parents, ensuring they are consistent and realistic, and provide a topological sampling recipe. \\
\\
Step 4: Noise Injection \\
You are a noise injector. \\
Given the clean SCM and a set of allowed noise types (VP, IV, PM, CS, CI, QP),
you must apply one or more noises, describe the modifications, and produce noisy variants. \\
\\
Step 5: QA Assembly \\
You are a data-to-text generator. \\
Given the SCM (clean or noisy) and semantic roles, you must generate a natural language background and formulate one reasoning question (observational/interventional/\\counterfactual) that requires causal reasoning.
\end{quote}

\paragraph{Part B: Graph-Guided Reasoning Framework}
\begin{quote}\ttfamily\small
Step 1: Parse Variables \\
You are a variable extractor. \\
Given the natural language background, you must identify all variables, their meanings, and conditional probabilities, and normalize them into structured form. \\
\\
Step 2: Verify Graph \\
You are a graph verifier. \\
Given the variable list and edges, you must check for acyclicity and consistency with the background, repair errors, and output a valid DAG with topological order. \\
\\
Step 3: Structure Input \\
You are an input organizer. \\
Given the graph, variables, and observed assignments, you must format the information into blocks: [Causal Graph], [Observed Variables], [Numbers]. \\
\\
Step 4: Reasoning \\
You are a causal reasoning engine. \\
Given the structured blocks and a formal query, you must compute the answer step by step, provide the derivation, and return the final numeric result.
\end{quote}